\documentclass[conference]{IEEEtran}

\usepackage{cite}

\usepackage[pdftex]{graphicx}
\graphicspath{{gfx/}}
\DeclareGraphicsExtensions{.pdf,.jpeg,.jpg,.png}

\usepackage{tikz}

\usepackage{amsmath}
\usepackage{array}
\usepackage{url}

\usepackage{adjustbox}
\newcommand{\includegraphicsfillbox}[3]
{\bgroup
  \dimen1=#1\relax
  \dimen2=#2\relax
  \sbox0{\includegraphics[width=#1]{#3}}%
  \ifdim\ht0>\dimen2
    \dimen0=\dimexpr \ht0-\dimen2\relax
    \adjustbox{clip=true,trim=0pt 0.5\dimen0 0pt 0.5\dimen0}{\usebox0}%
  \else
    \sbox0{\includegraphics[height=#2]{#3}}%
    \ifdim\wd0>\dimen1
      \dimen0=\dimexpr \wd0-\dimen1\relax
      \adjustbox{clip=true,trim=0.5\dimen0 0pt 0.5\dimen0 0pt}{\usebox0}%
    \else
      \usebox0
    \fi
  \fi
\egroup}

\usepackage{enumitem}
\usepackage{hyperref}
\usepackage{cleveref}
\usepackage{booktabs}
\usepackage{xcolor}
\usepackage{textcomp}

\usepackage{listings}

\makeatletter
\def\lst@makecaption{%
  \def\@captype{table}%
  \@makecaption
}
\makeatother

\AtBeginDocument{}

\crefname{lstlisting}{listing}{listings}
\Crefname{lstlisting}{Listing}{Listings}

\definecolor{delim}{RGB}{20,105,176}
\definecolor{numb}{RGB}{106, 109, 32}
\definecolor{string}{rgb}{0.64,0.08,0.08}

\lstdefinelanguage{json}{
    numbers=left,
    numberstyle=\small,
    frame=single,
    rulecolor=\color{black},
    showspaces=false,
    showtabs=false,
    breaklines=true,
    postbreak=\raisebox{0ex}[0ex][0ex]{\ensuremath{\color{gray}\hookrightarrow\space}},
    breakatwhitespace=true,
    basicstyle=\ttfamily\small,
    upquote=true,
    morestring=[b]",
    stringstyle=\color{string},
    literate=
     *{0}{{{\color{numb}0}}}{1}
      {1}{{{\color{numb}1}}}{1}
      {2}{{{\color{numb}2}}}{1}
      {3}{{{\color{numb}3}}}{1}
      {4}{{{\color{numb}4}}}{1}
      {5}{{{\color{numb}5}}}{1}
      {6}{{{\color{numb}6}}}{1}
      {7}{{{\color{numb}7}}}{1}
      {8}{{{\color{numb}8}}}{1}
      {9}{{{\color{numb}9}}}{1}
      {\{}{{{\color{delim}{\{}}}}{1}
      {\}}{{{\color{delim}{\}}}}}{1}
      {[}{{{\color{delim}{[}}}}{1}
      {]}{{{\color{delim}{]}}}}{1},
}

\newcommand\YAMLcolonstyle{\color{red}\mdseries}
\newcommand\YAMLkeystyle{\color{black}\bfseries}
\newcommand\YAMLvaluestyle{\color{blue}\mdseries}

\makeatletter

\newcommand\language@yaml{yaml}

\expandafter\expandafter\expandafter\lstdefinelanguage
\expandafter{\language@yaml}
{
  keywords={true,false,null,y,n},
  keywordstyle=\color{darkgray}\bfseries,
  basicstyle=\YAMLkeystyle,                                 
  sensitive=false,
  comment=[l]{\#},
  morecomment=[s]{/*}{*/},
  commentstyle=\color{purple}\ttfamily,
  stringstyle=\YAMLvaluestyle\ttfamily,
  moredelim=[l][\color{orange}]{\&},
  moredelim=[l][\color{magenta}]{*},
  moredelim=**[il][\YAMLcolonstyle{:}\YAMLvaluestyle]{:},   
  morestring=[b]',
  morestring=[b]",
  literate =    {---}{{\ProcessThreeDashes}}3
                {>}{{\textcolor{red}\textgreater}}1     
                {|}{{\textcolor{red}\textbar}}1 
                {\ -\ }{{\mdseries\ -\ }}3,
}

\lst@AddToHook{EveryLine}{\ifx\lst@language\language@yaml\YAMLkeystyle\fi}
\makeatother

\newcommand\ProcessThreeDashes{\llap{\color{cyan}\mdseries-{-}-}}

\usepackage[disable]{todonotes}

\usepackage{flushend}

\usepackage{background}

\newcommand\copyrighttext{%
  \footnotesize \textcopyright 2021 IEEE. Personal use of this material is%
  permitted. Permission from IEEE must be obtained for all other uses, in any%
  current or future media, including reprinting/republishing this material for%
  advertising or promotional purposes, creating new collective works, for resale%
  or redistribution to servers or lists, or reuse of any copyrighted component%
  of this work in other works. DOI:%
  \href{https://doi.org/10.1109/ETFA45728.2021.9613662}{10.1109/ETFA45728.2021.9613662}%
}

\newcommand\copyrightnotice{%
  \begin{tikzpicture}[remember picture,overlay]%
    \node[anchor=south,yshift=24pt] at (current page.south) {\fbox{\parbox{\dimexpr\textwidth-\fboxsep-\fboxrule\relax}{\copyrighttext}}};%
  \end{tikzpicture}%
}

\newcommand\makecopyrightnotice{%
  \backgroundsetup{opacity=1, scale=1, angle=0, color=black, contents={\copyrightnotice}}%
}

\IEEEoverridecommandlockouts
\IEEEpubid{
  \hspace{1mm}  
  \makebox[\columnwidth]{978-1-7281-2989-1/21/\$31.00~\copyright2021 IEEE \hfill}
  \hspace{\columnsep}
  \makebox[\columnwidth]{}
}

\begin{document}
%

\title{A Solution to the Generalized ROS Hardware IO Problem -- A Generic Modbus/TCP Device Driver for PLCs, Sensors and Actuators}

\author{%
  \IEEEauthorblockN{Arne Wendt}
    \IEEEauthorblockA{%
    \textit{Institute for Aircraft-Production-Technology}\\
    \textit{Hamburg University of Technology}\\
    Hamburg, Germany\\
    0000-0002-5782-3468%
    }
  \and
  \IEEEauthorblockN{Prof. Dr.-Ing. Thorsten Schüppstuhl}
    \IEEEauthorblockA{%
    \textit{Institute for Aircraft-Production-Technology}\\
    \textit{Hamburg University of Technology}\\
    Hamburg, Germany\\
    0000-0002-9616-3976%
    }
}

%


\maketitle

\makecopyrightnotice

\begin{abstract}
The Robot Operating System (ROS) provides a software framework, and ecosystem of
knowledge and community supplied resources to rapidly develop and prototype
intelligent robotics applications. By standardizing communication, configuration
and invocation of software modules, ROS facilitates reuse of device-driver and
algorithm implementations. Using existing implementations of functionality allows
users to assemble their robotics application from tested and known-good
\emph{capabilities}. Despite the efforts of the ROS-Industrial consortium and
projects like ROSIN to bring ROS to industrial applications and integrate
industrial hardware, we observe a lack of options to generically integrate basic
physical IO. In this work we lay out and provide a solution to this problem by
implementing a generic Modbus/TCP device driver for ROS.
\end{abstract}


%
\IEEEpeerreviewmaketitle

\section{Introduction}\label{intro}
The Robot Operating System (ROS) is a meta-operating system for robots. It shall
provide users with "hardware abstraction, low-level device control,
implementation of commonly-used functionality, message-passing between
processes, and package management"\cite{rosintro}. Supplied by either ROS itself,
or as community supplied resources. By building an ecosystem of knowledge and
supplemental community supplied software resources, it allows to rapidly
assemble and prototype intelligent robotics applications.
By standardizing communication, configuration and invocation of software
modules, ROS facilitates reuse of device-drivers, algorithm implementations and
tooling. Using existing implementations allow users to assemble their robotics
application from tested and known-good \emph{capabilities}.

ROS has a strong focus on service-robotics and intelligent autonomous mobile
robotics application, providing out-of-the-box solutions to common high level
problems like e.g. path-planning, navigation and execution of (mostly) wheeled
locomotion, using the \emph{navigation stack}\cite{navstack}, as well as
trajectory planning and execution for multi-axis manipulators and grippers using
\emph{MoveIT}\cite{moveit}. These high level solutions are complemented by
sensor-drivers, enabling the necessary perception for the planning tasks and
subsequent execution. Founded In 2012, and supported by projects like ROSIN
\cite{rosin}, the ROS-Industrial\cite{rosindustrial} consortium aims to bring
ROS to industrial applications and integrate industrial hardware into the ROS
ecosystem.
Industrial applications, especially automation in production and handling, are
starkly characterized by their interaction with and physical manipulation of the
real world. Thus, for a successful application of ROS in industrial application,
there exists a strong demand for integration of actuators and sensors of
different modalities.

Even tough these components are readily available, and usually implement well
defined and standardized interfaces, we observe a lack of options to integrate
this wealth of existing sensors and actuators from (not limited to, but
especially) industrial applications into the ROS environment in a generic way.
This deficiency is as far ranging, as the integration of a simple button or
switch poses a serious hurdle in ROS application development. To our knowledge,
no solution to this broader problem, providing IO integration with sufficient
scalability, portability and flexibility, exist within the ROS ecosystem today.

With the following work we want to show a way, and provide the
accompanying tools (software), we deem an acceptable solution to the
problem stated above; in industrial, as well as research and hobbyist
applications.

\IEEEpubidadjcol

\section{Hardware Drivers in ROS}
\label{drivers}

Reviewing the official ROS community resources \cite{roswiki} and
\cite{rosdiscourse}, the ROS industrial resources \cite{rosindustrialsupported},
as well as publicly available projects on \cite{rosdrivers}, we identify
\textbf{\emph{five}} \emph{main} categories of drivers:
\footnote{We abstain from providing a (exemplary) list of reviewed driver,
  as we are neither conducting a systematic review, nor do we believe that a -
  under all circumstances non-exhaustive - list would be of particular benefit
  to the reader}

\begin{enumerate}
\item
  2D vision sensors (cameras)
\item
  1D - 3D depth sensors: mostly LIDaR, ToF and stereo-vision
\item
  Robotic arms/manipulators
\item
  Vendor specific "robotics" platforms
\item
  \emph{non-ROS-enabled} wrappers for (field bus) communication protocol
  implementations
\end{enumerate}

The last category mostly just wraps libraries for ROSs' dependency management
and build system (\emph{catkin}) and is of no further interest here.

We can further classify the first \textbf{\emph{four}} driver categories
by the following properties:

\begin{enumerate}[label=\Alph*]
\item
  well defined hardware-interfaces with same
  \textbf{semantic} output (data class) and small
  variable-parameter-vector
\item
  high market penetration of targeted hardware, or low availability of
  competitive products
\end{enumerate}

Where an example for category \textbf{A} would be the \texttt{usb\_cam}
driver\cite{usbcam}, providing an interface from \texttt{v4l}/\texttt{v4l2}
(\textbf{V}ideo \textbf{for} \textbf{L}inux (2)) compatible cameras to ROSs'
standardized \texttt{sensor\_msgs/Image} format. Category \textbf{B} would be covered by
(mostly) vendor specific drivers for e.g. LiDAR-Scanners, like
\textbf{\emph{SICK}}(cmp. \cite{sick}), or \textbf{\emph{Intel Realsense}} depth
cameras (cmp. \cite{realsense}). The two categories are non-exclusive though:
Common drivers for each (most) all e.g. \textbf{\emph{ABB}} or
\textbf{\emph{Universal Robots}} industrial robotic arms (cmp. \cite{abb,ur}),
or \textbf{\emph{iDS uEye}} cameras (cmp. \cite{ueye}), fit both categories.

The first category poses a high incentive for the community and
community developed drivers, as standardized interfaces usually increase
the available options and this decrease the price; ultimately increasing
availability and accessibility. The second category allows for two
driving mechanism in driver development: First, manufacturers targeting
the ROS community as a market. Second, a demand for specific sensor
modality or properties, driven by the community.

ROS has ever been targeting *NIX based computer platforms in general,
and the Ubuntu platform in particular\cite{rosrep3}. It is thus
not surprising to observe another common property across the available
ROS drivers:

\begin{enumerate}[label=\Alph*,resume]
  \item
  ROS drivers target Hardware with interfaces commonly found on general purpose
  computing platforms, e.g Universal Serial Bus (USB), serial COM interfaces
  (RS232/RS485/UART) or Ethernet connections.
\end{enumerate}

\section{Problem Statement}

\emph{Basic} industrial IO components (sensors and actuators) usually expose
their functionality as one input or output per individual function; either
logic/digital (on/off) or analog (continuous range of values). These interfaces
adhere to the basic specifications of $\{5V, 10V, 24V\}$ for digital and logic
IO, and voltage and current signal in the ranges of $[0V, 10V]$ and $[4mA,
20mA]$ for analog signals as defined by \cite{iecvolt,ieccurr}.

When comparing the above mentioned interfaces to property
\textbf{C} of available ROS drivers, we can immediately observe a
problem:

\begin{quote}
\emph{Recent general purpose computing platforms do not provide individual
connections for general purpose logic or analog signal inputs.}
\end{quote}

For common IO components or simple applications like switch/contact
readout, we can further imagine property \textbf{B} to not be
fulfilled. Applying the abstraction of all functionality to the two
generic interface modalities, logic IO for discrete binary states, and
analog IO for the representation of a closed continuous interval of
values, it is trivial to see property \textbf{A} to not apply in
this case as well.

The lack of physical interfaces on current computing platforms poses a serious
technical hurdle to the implementation of ROS drivers for sensors and actuators;
without additional hardware and specifications, an integration is impossible.
The lack of a ubiquitously available hardware interface, and knowing property
\textbf{A} to not apply, further lowers the interest of the community to drive a
generic driver development.

\begin{figure}[!t]
  \centering
  \includegraphics[width=\linewidth]{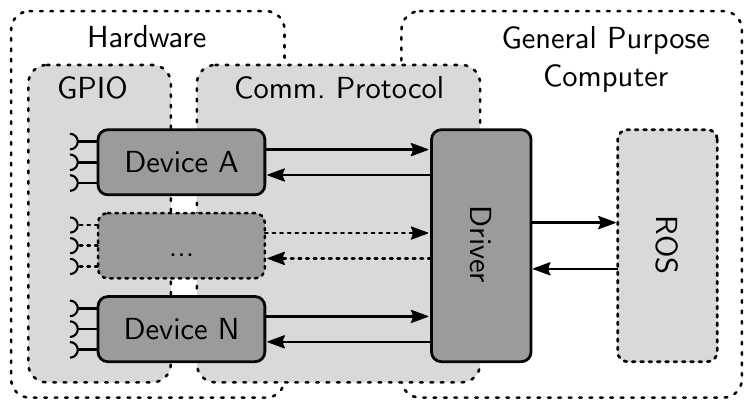}
  \caption{Schematics of information flow from hardware IO to ROS, showing
  components and properties to identify, as well as their relations and
  interfaces; class or set of GPIO capable hardware, as \emph{Device A} to
  \emph{Device N}}
  \label{pipeline}
\end{figure}

To provide a solution to the \emph{generalized problem} of a flexible, portable
and scalable \emph{hardware IO} integration into the ROS ecosystem, we first
need to identify a set or class of possible hardware interfaces providing the
necessary general purpose IO (GPIO) capabilities, a common communications
protocol to mediate between hardware and software layer, and subsequently define
and implement a generalized software abstraction on top of its concepts. The
general structure of which is show in \cref{pipeline}

\section{Choosing a Hardware Interface and Communications Protocol}\label{choosing}

Providing general purpose IO capability and transferring acquired data or
accepting setpoint values over a standardized communications protocol is a
common application in industrial automation system. Vendors like e.g.
\emph{Beckhoff}, \emph{Wago} and \emph{Phoenix Contact}, provide bus couplers
for all major industrial communication protocols and busses. These devices
themselves are then expandable in their IO capabilities by adding additional
IO-modules; the same applies to PLCs. Bus coupler and PLC manufacturers usually
provide a wealth of IO modules for their respective platform, enabling
sufficiently \emph{general purpose} IO in respect to the problem at hand.
Consequently, for the remaining part of the work, we will assume the
device-class of bus couplers to be the best representation of a solution to the
hardware-facing portion of the overall problem; mapping GPIOs to a standardized
protocol. \emph{Further evaluation and selections will be made with bus couplers
as the primarily targeted hardware.}



To provide an as generic as possible solution to the generalized problem, we
want the following requirements to be met:

\emph{Scalability:}
\begin{itemize}
\item
  Support hardware interfaces with an extendable IO count; enabling
  \emph{vertical} scalability
\item
  Support a scalable number of small (cheap) devices to increase IO count;
  enabling \emph{horizontal} scalability
\end{itemize}

\emph{Portability:}
\begin{itemize}
\item
  Not target one specific device
\item
  Not target vendor-specific hardware or communication protocols
\item
  Prefer technical solutions without the need for specialized
  communications hardware
\item
  Be hardware-agnostic regarding the hardware platform running ROS
\item
  Shall use commonly available hardware interfaces on the hardware
  platform running ROS, e.g. USB, serial COM, Ethernet or WiFi
\end{itemize}

\emph{Flexibility:}
\begin{itemize}
\item
  Provide a single integrated conceptual solution to generic logic and
  analog IO
\item
  Allow for mapping of hardware/protocol data to
  basic ROS message types and topic trees
\end{itemize}

\subsection{Communications Protocol}


Bus couplers and PLCs usually implement fieldbus interfaces. The resulting
challenge is the identification of an interface to be used in combination with
the hardware running ROS. As analyzed by \cite{industrialcomm}, recent
developments of fieldbuses - and communication protocols in industrial
applications in general - show an increase in the use of ethernet media access
(MAC) and physical layer (PHY). Reviewing the current bus coupler offerings of
major manufacturers (as above in \cref{choosing}) we support these findings,
while simultaneously identifying Ethernet and RS232/RS485 based solutions as the
only commonly available protocols with a suitable PHY. As manufacturers are
increasingly adopting ethernet based solution, and serial COM ports are on the
decline on recent computer hardware, we rule out the use of serial COM protocols
in favor of ethernet based solutions.

The resulting list of commonly available and standardized communication
protocols using an ethernet PHY, providing cross-vendor-compatibility, is given
as column one of \cref{tableeth}.
Analyzing the properties of the potential candidates, with the help of
\cite{ethernet,industrialcomm,opcrosmqtt,efficientiot}, we can further compare
their benefits and deficiencies and select an optimal solution to our problem.
The compared properties and their rating will be made in reference to the
satisfaction of our requirements. We identify the following parameters and their
associated impacts as a first set of filters relevant to our problem and
requirements:

\begin{itemize}
\item
  \emph{Maximum number of devices per protocol}; impacts scalability and
  portability to existing environments
\item
  \emph{Requirement for special hardware}; impacts portability
  requirements
\end{itemize}

\begin{table}[!t]
  \renewcommand{\arraystretch}{1.3}
  \caption{
    Maximum Device Number and Special Hardware Requirements Indication for
    Ethernet Based Communication Protocols (partly based on \cite{ethernet})
    }
  \label{tableeth}
  \centering
  \begin{tabular}{lrr}
    \toprule
    Protocol    & Max. no. Devices & Special Hardware Required \\
    \midrule
    Profinet    & 60            & yes                       \\
    EtherNet/IP & 90            & yes                       \\
    Modbus/TCP  & -             & no                        \\
    EtherCAT    & 180           & yes                       \\
    OPC UA      & -             & no                        \\
    MQTT        & -             & no                        \\
    \bottomrule
  \end{tabular}
\end{table}

With the results, as given in \cref{tableeth}, we can further narrow the
selection to the three solutions not capping the maximum device number
and not requiring special hardware for operations.

\subsubsection{Protocol Overview}
Properties of the three remaining contenders will be described briefly;
focusing on conceptual differences and feature set, as well as a short analysis
on the availability of protocol implementations in suitable hardware devices.

\subsubsection*{OPC UA}
The OPC Unified Architecture defines a layered architecture for communications
between devices \cite{opcua1}. While the specification mostly defines concepts and behaviors,
standardized mappings (\cite{opcua6}) to concrete technical implementations exist and provide
the flexibility to choose the application specific optimal technologies. At its
core, OPC UA specifies a methodology for information modelling, specified in \cite{opcua5}. With this
information model, OPC UA allows to describe, among other information, the
available data/variables to be read from and written to a server (device),
modelling e.g. hierarchical relationships and annotating values with 
datatype, physical unit and a (unique) name/identifier. Providing a descriptive
information model of the provided data is a highly desirable trait of the OPC UA
communication stack, as it allows for automatic and dynamic configuration of a
driver (or client in general) from the devices' information model.

Implementations of OPC UA servers are commonly available in recent PLCs.
Reviewing the functionality of implementations provided by \emph{Wago},
\emph{Beckhoff} and \emph{Siemens}, we find that only variables present in a PLC
program may be exposed over OPC UA. No direct access to IO modules attached to
the PLC is possible via OPC UA without mapping these to a variable. While OPC UA
server support in PLCs seems in place, reviewing the OPC foundations' list of
certified products\cite{opcfind}, we identify only a single
device\footnote{\emph{Beckhoff EK9160}} implementing bus coupler like behavior
and exposing its attached IO modules directly via OPC UA.

\subsubsection*{MQTT}

The Message Queuing Telemetry Transport (MQTT)\cite{mqtt} provides means of
network communication built on the TCP/IP stack, targeted at IoT applications.
MQTT is a pure transport and messaging implementation and does not define
payload data serialization. Without a standardized and widely adopted
serialization, MQTT alone cannot ever satisfy our requirement for flexibility.
Sparkplug\cite{sparkplug} is a specification for topic organization of the MQTT
publish-subscribe-architecture and serialization of MQTT payloads, intended to
provide compatibility of devices and applications in industrial IoT (IIoT)
applications. Sparkplug provides means to discover device capabilities and
provided data, as well as device/node discovery on the network.

While MQTT alone and in conjunction with Sparkplug seem to provide a desirable
performance over competing protocols (cmp. \cite{opcrosmqtt,efficientiot}) we
are unable to identify a serious\footnote{To the best of our knowledge, only two
product lines of PLCs and one bus coupler do natively support MQTT and
Sparkplug; namely \emph{groov EPIC} and \emph{groov RIO} by \emph{Opto 22} and
the \emph{Wago PFC200 series}} number of devices implementing Sparkplug.

\emph{For the remainder of this work, MQTT will be used to refer to the
combination of MQTT and Sparkplug.}

\subsubsection*{Modbus/TCP}

Modbus/TCP, at its core, allows to call defined methods on a remote device and
receive the corresponding function call result. Methods for data access build on
the Modbus data model; both specified by \cite{modbus}. This basic data model
distinguishes between bit and register values, while methods for read and write
operation for each singular or multiple of these values are provided. While
Modbus does not provide any standardized means to communicate meta information
like data type or physical unit, Modbus/TCP
enabled devices commonly adhere to the data types specified by
\emph{IEC-61131-3}\cite{iecplc}.

Apart from specialized Modbus/TCP-bus couplers, virtually all PLCs and bus
couplers with an ethernet connection provide a Modbus/TCP server/slave
implementation. Modern PLCs using a \emph{IEC-61131-3} compatible runtime,
without a vendor supplied server, may add Modbus/TCP compatibility as a software
package/module\cite{codesysmodbus}.

\subsubsection{Protocol Selection}

As all three protocols are publicly available and open standards, and are either
based on -- or provide capability to leverage -- standard ethernet based
communications, we assume all three to directly satisfy all but our first
requirements in regards to \emph{portability}. Further, all protocols do support
transmission of individual data points and types for representation of analog
and digital IO, thus satisfying our requirements in relation to
\emph{flexibility}.

MQTT and OPC UA, each seem to provide technically optimal solution to the
integration and exchange of IO information with hardware acquisition devices,
for the problem at hand. Both provide bidirectional data transmission, either
explicit or implicit discovery mechanisms, and data typing. The goal of
providing an actually \textbf{applicable} solution, prohibits the evaluation of
communication protocols independently of the availability of suitable hardware.
The fulfillment of our requirements relating to scalability is tightly coupled
to hardware availability, and cost/simplicity of protocol implementation on
devices to enable horizontal scalability.

Based on the analysis of available implementations in hardware, we conclude that
neither OPC UA nor MQTT are sufficiently supported by available hardware. While
OPC UA has good support on PLCs, support in bus couplers -- as object of
consideration (compare \cref{choosing}) -- is negligible. The overall support of
MQTT is considered negligible as well. The low number of hardware options
supporting \emph{vertical scalability}, makes both OPC UA and MQTT not fulfill
our first requirement of \emph{portability} and thus effectively not satisfying
the requirement of \emph{vertical scalability} itself. Modbus/TCP, on the other
hand, constitutes a technically inferior solution, while showing significantly
higher adoption and availability of supported hardware, thus fulfilling the
requirements of \emph{portability} and \emph{vertical scalability}. The fitness
for \emph{horizontal scalability}, usually enabled by low cost -- commonly
implying low resources -- of supporting devices, can be assumed to be inversely
proportional to the complexity of a technology. Assessing each protocols
complexity in comparison to each other, we can establish following order from
least complex to most: Modbus/TCP, MQTT, OPC UA. Using this scale as an
indicator for \emph{horizontal scalability}, Modbus/TCP constitutes the
preferred choice.

\emph{Based on the above evaluation, we identify \emph{Modbus/TCP} as the
\emph{currently} optimal solution to our problem, and choose it as the common
denominator for integrating hardware IO with current general purpose computing
hardware running ROS.}

As its core functionality, Modbus/TCP allows read and write access to memory
regions -- usually representing IO states or program variables -- on a device.
While not specifying data types and being able to deduce the data type from
register data itself, most Modbus/TCP enabled devices adhere to the data types
specified by \emph{IEC-61131-3}. While the nature of Modbus/TCP requires the
specification of addresses to be polled for data, thus demanding additional
efforts for configuration, this property allows to map these individually
requested values to ROS topics with a specified data type.\todo{absatz evaluieren}

Being an old protocol, in comparison to its contestants, Modbus/TCP still has
wide support - not limited to industrial applications. Internet of Things (IoT)
cloud solution providers like Microsoft Azure \cite{azure}, AWS IoT SiteWise
\cite{aws} or Alibaba Link IoT Edge \cite{alibaba}, support data integration for
Modbus/TCP devices.

\subsection{Hardware Compatibility Consideration}

Our requirements for portability forbid the selection of specific hardware
devices. Consequently the \emph{hardware interface} is represented by a class of
devices implementing our chosen communications protocol.

The simplicity of the Modbus/TCP protocol and its foundation on the TCP/IP
protocol stack allow for good compatibility across a wide range of platforms. As
support and availability of implementations in PLCs and bus couplers has been
the driving factor in the protocol selection, compatibility for these classes of
devices is granted.

For horizontal scalability, low cost and power IoT devices and hardware
platforms provide a solution, while receiving good support for Modbus/TCP
compatibility. This class of devices is targeted by operating systems like
FreeRTOS\cite{freertos} with available ports of Modbus/TCP clients like
FreeMODBUS\cite{freemodbus}, as implemented e.g. by the ESP-IDF (Espressif IoT
Development Framework)\cite{espidf} for the ESP32 System on Chip platform.
Bringing Modbus/TCP capability to these low power and low cost microcontroller
platforms, increases availability and allows horizontal scalability through low
cost.

Generally, all device and operating system combinations providing a TCP/IP
protocol stack are able to serve as the base for a Modbus/TCP IO device.

\section{Implementation}

Despite the listing of Modbus/TCP as a hardware interface protocol in the
proposed ROS Industrial roadmap, since at least 2015
\cite{rosindustrialroadmapgithub,rosindustrialroadmap}, there exists no
generic implementation of a driver for the ROS ecosystem.

To provide a solution to the generalized IO problem, based on Modbus/TCP, we
choose to implement a driver with the following major properties:
\begin{itemize}
  \item each driver-node instance represents each one slave device or subset of
  a slaves' mapped IOs
  \item each driver with associated mapping uses an individual topic namespace
  \item map each configured IO to a separate topic
  \item support read access for \emph{discrete inputs} and \emph{input
  registers}(\emph{status registers})
  \item support read/write access for \emph{coils} and \emph{output
  registers}(\emph{holding registers})
  \item perform data conversion for register data from \emph{IEC 61131-3} to ROS
  \texttt{std\_msgs} types (or arrays thereof) and vice versa
\end{itemize}

The first three properties achieve a separation of concerns in functionality and
allow to follow ROS' ideas of reusability and composition. The remaining
properties establish a sufficient base functionality for a generic driver in the
targeted domain. Mapping IOs to their own topics each is the best representation
of our generic driver approach, while enabling integration of IOs via ROSs'
topic remapping functionality.

\begin{figure}[!t]
  \centering
  \includegraphics[width=\linewidth]{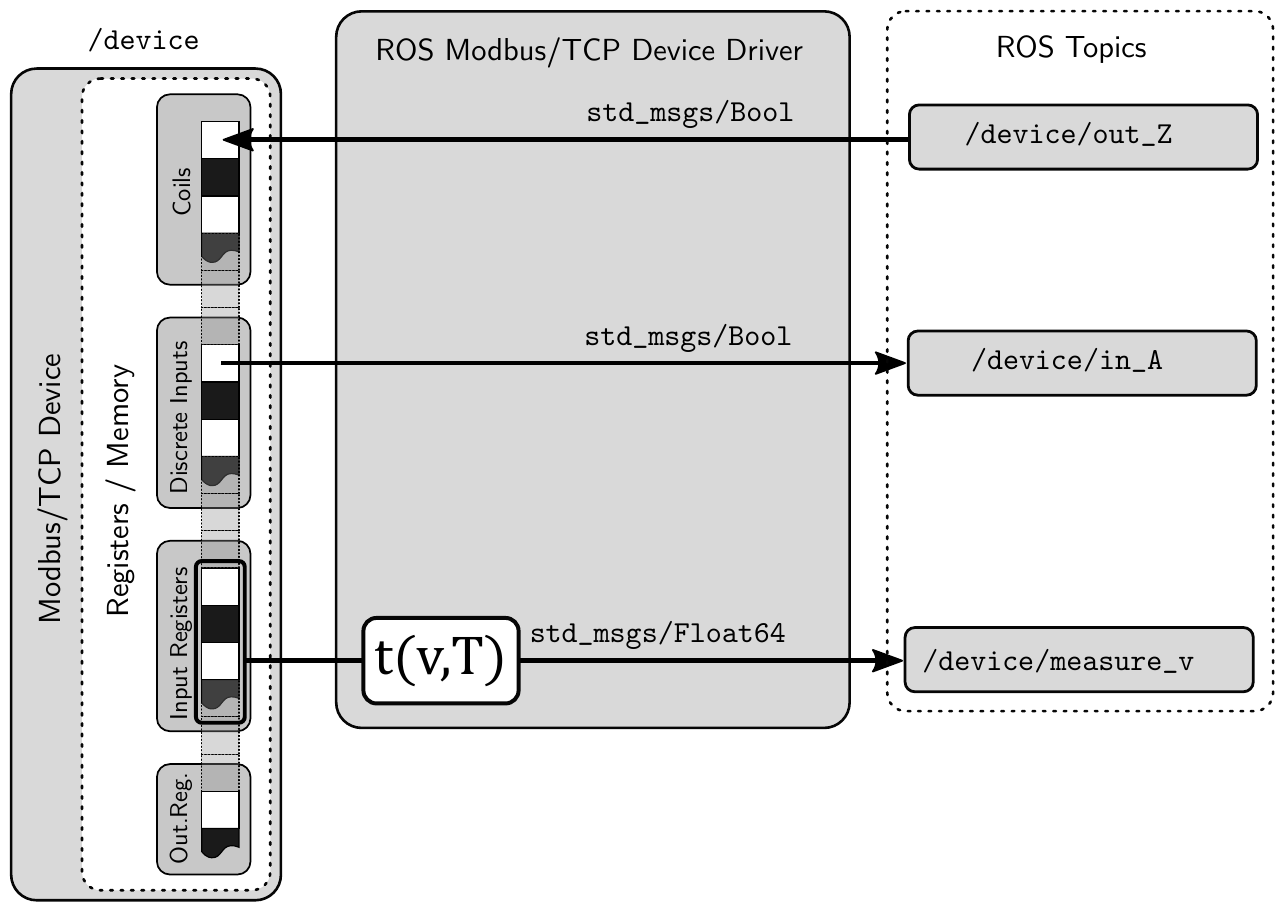}
  \caption{
    Basic ROS Modbus/TCP driver principle. Mapping Modbus/TCP items to
    individual ROS topics. Data interpretation from registers show as
    conversion-method $t$, converting from register value $v$ to ROS compatible
    representation with data type $T$.
    }
  \label{modbusros}
\end{figure}

\cref{modbusros} shows an example of the drivers functionality. We map a
Modbus/TCP device to ROS namespace \texttt{/device}, a \emph{coil} (logic
output) as \texttt{/out\_Z}, a \emph{discrete input} as \texttt{in\_A} and a 4
word (8 bytes) wide \emph{input register} as \texttt{/measure\_v}. The 4 words
of the \emph{input register} are interpreted as \emph{IEC 61131-3}
\texttt{LREAL} (\emph{long real}, $64 bit$ floating point number), corresponding
to ROSs' \texttt{std\_msgs/Float64}. The driver polls the slave device at a
constant rate and publishes the acquired values to the mapped
topics. On reception of messages on a topic mapping an output, the received
value will be transformed if necessary, by applying the inverse of $t$ (compare
\cref{modbusros}); compatible and driver-supported (implemented) \emph{IEC
61131-3} types and corresponding ROS \texttt{std\_msgs}
types are shown in \cref{typetab}.

\begin{table}[!t]
  \renewcommand{\arraystretch}{1.3}
  \caption{\emph{IEC 61131-3} types to ROS {\normalfont\texttt{std\_msgs}} mapping}
  \label{typetab}
  \centering
  \begin{tabular}{llc|cll}
      \toprule
      \emph{IEC 61131-3} & \texttt{std\_msgs/}&\multicolumn{2}{c}{}& \emph{IEC 61131-3} & \texttt{std\_msgs/}\\
      \cmidrule{1-2}\cmidrule{5-6}
      \texttt{BYTE}    & \texttt{Bool[8]}     &\multicolumn{2}{c}{}& \texttt{DWORD}   & \texttt{Bool[32]}\\
      \texttt{WORD}    & \texttt{Bool[16]}    &\multicolumn{2}{c}{}& \texttt{LWORD}   & \texttt{Bool[64]}\\
      \cmidrule{1-2}\cmidrule{5-6}
      \texttt{SINT}    & \texttt{Int8}        &\multicolumn{2}{c}{}& \texttt{DINT}    & \texttt{Int32}\\
      \texttt{INT}     & \texttt{Int16}       &\multicolumn{2}{c}{}& \texttt{LINT}    & \texttt{Int64}\\
      \cmidrule{1-2}\cmidrule{5-6}
      \texttt{USINT}   & \texttt{UInt8}       &\multicolumn{2}{c}{}& \texttt{UDINT}   & \texttt{UInt32}\\
      \texttt{UINT}    & \texttt{UInt16}      &\multicolumn{2}{c}{}& \texttt{ULINT}   & \texttt{UInt64}\\
      \cmidrule{1-2}\cmidrule{5-6}
      \texttt{REAL}    & \texttt{Float32}     & & & \texttt{CHAR}    & \texttt{Char}\\
      \texttt{LREAL}   & \texttt{Float64}     & & & \texttt{STRING}  & \texttt{String}\\
      \bottomrule
  \end{tabular}
\end{table}

\subsection{ROS to Modbus/TCP interface}
\begin{figure}[!t]
  \centering
  \includegraphics[width=\linewidth]{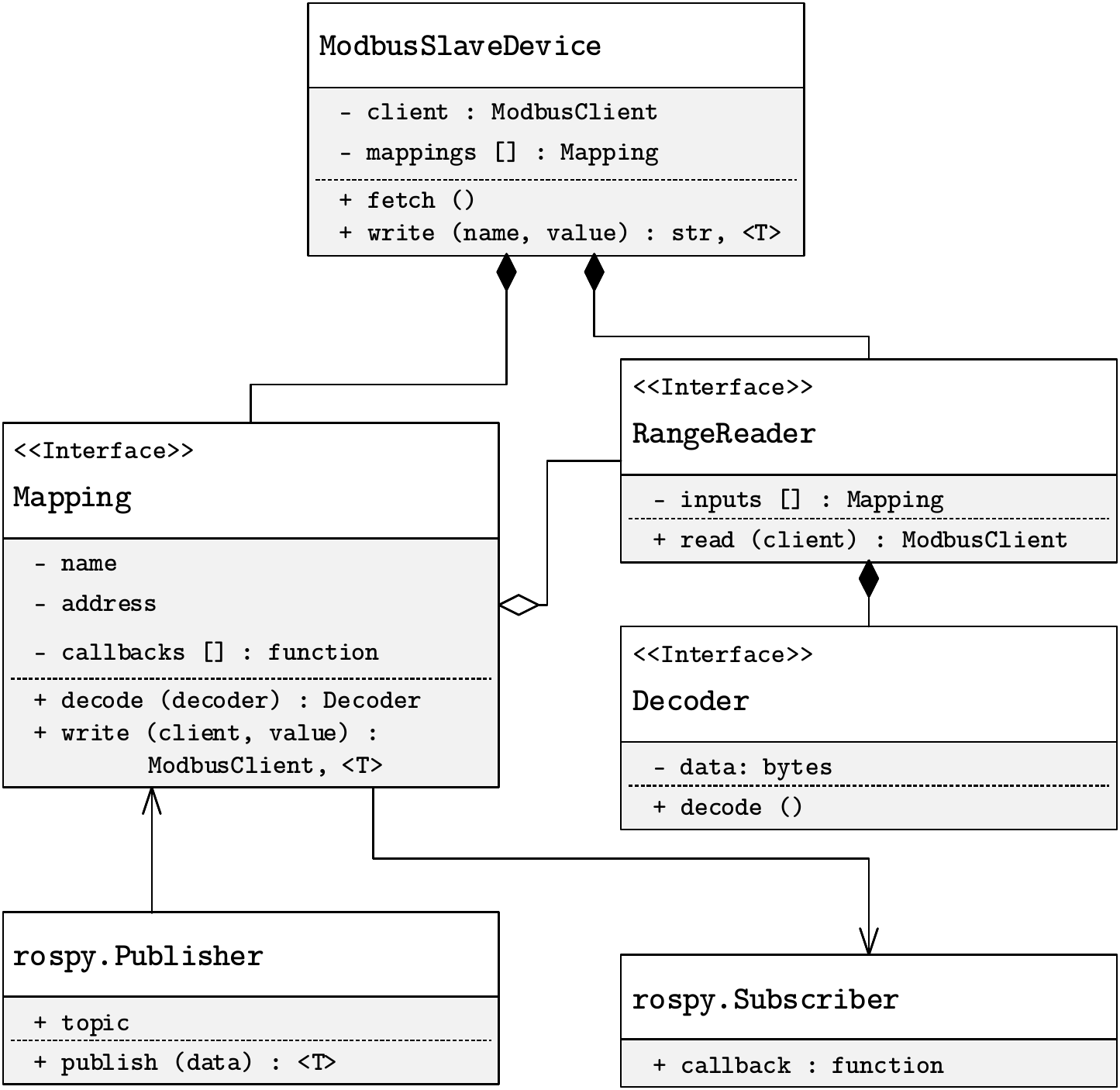}
  \caption{
    Simplified and abstracted class relationship diagram of the Modbus/TCP slave
    device wrapper \texttt{ModbusSlaveDevice}.  
    }
  \label{classdiagram}
\end{figure}
Following, we will give an overview of how the data acquisition process from a
slave device to ROS topics, and the process of writing data from a topic to a
slave is implemented. For this we will use the simplified and abstracted class
relationship diagram from \cref{classdiagram}, omitting intermediate classes
responsible for state management and implementing behaviors not relevant to the
overall operation principle. We further want to assume all behavioral
differences to be implemented by class inheritance and specialization, while
implementing the shown interfaces in \cref{classdiagram}. We begin with a brief
description of the non-ROS classes, followed by a description of the process for
reading and writing data.\\


The driver itself provides an abstraction over one Modbus/TCP device with a
selection of mapped IOs. The ROS node holds and manages an instance of a
\texttt{ModbusSlaveDevice}, provides configuration and manages ROS publishers
und subscriptions (\texttt{rospy.Publisher} and \texttt{rospy.Subscriber} in
\cref{classdiagram}).


\paragraph*{\texttt{ModbusSlaveDevice}} Abstracts a Modbus/TCP device. Manages and
persists a connection to the Modbus/TCP slave, holds class instances of all
mappings from Modbus addresses to ROS topics and associated \texttt{RangeReader}s.
The \texttt{fetch()} method triggers read operations for all \texttt{RangeReader}s.

\paragraph*{\texttt{Mapping}} Each input and/or output and its associated topics
are represented by a class instance implementing the \texttt{Mapping} interface.
Mappings are (for the sake of simplicity, assumed to be) specialized on
\emph{discrete inputs}, \emph{input registers}, \emph{coils} and \emph{output
registers}, as well as the type of register data. The \texttt{decode()} method
will fetch the registers' value from a \texttt{Decoder} and call all associated
\texttt{callbacks} with the decoded value. Calling \texttt{write()} method will
encode the passed data according to type and instantly trigger a write operation
to the Modbus/TCP slave.

\paragraph*{\texttt{RangeReader}} Aggregates read operations for multiple
\texttt{Mapping}s to one read operation of a continuous memory region, reducing
polling operations and network traffic. Depending on slave behavior, reads from
discontinuous memory regions can be aggregated while discarding unmapped memory.

\paragraph*{\texttt{Decoder}} Container, owning and representing the raw data
returned by a \texttt{RangeReader}s' \texttt{read()} operation. Implements
data decoding functionality from raw data to requested type. \texttt{Mapping}s
decode their value from the \texttt{Decoder} depending on type.

\subsubsection{Writing Outputs}
The driver node attaches the respective \texttt{Mapping}s' \texttt{write()} method
as callback function to a ROS subscriber (\texttt{rospy.Subscriber}) instance for
the associated topic. The \texttt{Mapping} class specialization handles encoding
of the data in relation to its type and triggers a write operation on the provided
\texttt{ModbusClient} instance.

\subsubsection{Reading/Polling Inputs}
Polling a slave is performed by the driver node by calling the
\texttt{ModbusSlaveClient}s' \texttt{fetch()} method. This in
turn triggers the individual processes described above: Triggering all
\texttt{read()} methods of the \texttt{RangeReader}s, which will start the
decoding process for all mappings on data reception, in turn leading to a call of the
\texttt{publish()} method of the respective \texttt{rospy.Publisher}.

\subsection{Configuration}
Creating a device abstraction and mapping its hardware IOs to ROS topics, is
accomplished by defining the abstraction as device configuration. The
configuration may be supplied in ROSs' standard configuration file format
\emph{YAML}, or \emph{JSON}. Assuming a slave with hostname
\texttt{mbdev}, Modbus unit \texttt{1} and matching IO addresses,
we can create a device mapping, representing the example in \cref{modbusros} as
in \cref{config}.\todo{adjust spacing with listing}

\begin{lstlisting}[
  float=t,%
  floatplacement=t,%
  language=yaml,%
  label=config,%
  numbers=none,%
  frame=tb,%
  basicstyle=\small\ttfamily,%
  caption={
    Example device mapping, corresponding to the example depicted in \cref{modbusros}
  }]
name: device

address: mbdev
unit: 1

rate: 20

mapping:
  coils:
    out_Z: 1

  discrete_inputs:
    in_A: 10001

  input_registers:
    measure_v:
      address: 30001
      type: LREAL
\end{lstlisting}

\vspace{1.25\baselineskip}
\emph{The complete drivers' implementation can be acquired and studied at
\cite{rosmodbusdevicedriver}.}

\section{Performance Analysis}

Translating from ROS messages to Modbus/TCP and vice versa, we will use the time
difference between TCP-packets, for ROS messages to Modbus/TCP write-request,
and Modbus/TCP read-responses to ROS messages, being available on the network as
the relevant performance metric. Read- and write requests from the driver node
will be handled by a dummy Modbus/TCP slave, accepting all request and
responding with a success-responses, returning $0$-values in case of read
operations. All data is generated and captured in its own docker container,
transferring data over the container internal loopback interface. Each container
is running the complete software stack necessary for data acquisition, where $n$
is the number of configured inputs or outputs for the case examined:
\begin{itemize}
  \item $1$ local \emph{roscore}
  \item $1$ driver node
  \item $n$ \emph{rostopic} nodes as publisher or subscribers
  \item $1$ \emph{Diagslave}\cite{diagslave} Modbus/TCP slave simulator
  \item $1$ \emph{tshark} (\emph{Wireshark}) instance for data capture
\end{itemize}
This test setup allows to capture measures for the actual delay introduced by
the driver and the ROS stack, while being independent of network induced delays
and the performance of the used Modbus/TCP slave.

Reading and/or writing occurs at a frequency of $10Hz$, while capturing network
traffic for $100s$, resulting in each $1000$ individual read or write operations
to be evaluated. All containers are ran with a limit of
$1GB$ RAM and $2$ CPUs, running on a \emph{Windows 10 Pro} host machine with an
\emph{Intel\textsuperscript{\textregistered}
Core\textsuperscript{\texttrademark} i7-8550U @} $1.80GHz$ CPU and $8GB$ of RAM.

\subsection{Reading}\label{performancereading} To evaluate read-performance, the
driver node is configured for either one or 32 \emph{discrete inputs} or
\emph{input registers}. For \emph{input registers}, interpretation of a
registers' value as a 16 bit integer value is chosen. Reading \emph{coils} or
\emph{holding registers} uses the same implementation as \emph{discrete inputs}
or \emph{input registers} and is thus not evaluated individually. For all inputs
each one \emph{rostopic} subscriber is instantiated. The average time difference
between Modbus/TCP read-responses and first (or only) ROS-message
$\overline{\Delta t_{0}}$ and average time difference between subsequent
ROS-messages $\overline{\Delta t_{P}}$, for each single- and 32-value read
operations, is given in \cref{dtdiscrete} for \emph{discrete inputs} and
\cref{dtinreg} for \emph{input registers}.

\begin{table}[!t]
  \renewcommand{\arraystretch}{1.3}
  \caption{Performance evaluation results for \emph{discrete inputs} as
  described in \cref{performancereading}}
  \label{dtdiscrete}
  \centering
  \begin{tabular}{rcc}
      \toprule
      No. Inputs & $\overline{\Delta t_{0}}$ & $\overline{\Delta t_{P}}$ \\
      \midrule
      1   & $341\mu s \,(\sigma=\hphantom{1}94\mu s)$   & -\\
      32  & $471\mu s \,(\sigma=192\mu s)$  & $182\mu s \,(\sigma=9\mu s)$\\
      \bottomrule
  \end{tabular}
\end{table}

\begin{table}[!t]
  \renewcommand{\arraystretch}{1.3}
  \caption{Performance evaluation results for \emph{input registers} as
  described in \cref{performancereading}}
  \label{dtinreg}
  \centering
  \begin{tabular}{rcc}
      \toprule
      No. Inputs & $\overline{\Delta t_{0}}$ & $\overline{\Delta t_{P}}$ \\
      \midrule
      1   & $388\mu s \,(\sigma=\hphantom{2}86\mu s)$   & -\\
      32  & $522\mu s \,(\sigma=203\mu s)$  & $187\mu s \,(\sigma=10\mu s)$\\
      \bottomrule
  \end{tabular}
\end{table}

\subsection{Writing}\label{performancewriting} Publishing to the \texttt{/write}
topic associated with an output will directly invoke a write request to the
Modbus/TCP slave; thus, only single value writes are technically possible and
evaluated. For each evaluation, a \emph{rostopic} publisher is instantiated,
writing the value $0$ (or the boolean equivalent \texttt{false}). For writing
\emph{holding registers} \texttt{std\_msg/Int16} is chosen as datatype, as in
\cref{performancereading}. Evaluating write performance, analogously to
\cref{performancereading}, $\overline{\Delta t_{0}}$ represents the time
difference between a ROS-message and the associated Modbus/TCP write-request
being available on the network.

Experiments yield a $\overline{\Delta t_{0}}=158\mu s (\sigma=17\mu s)$ for
writing \emph{coils}, and $\overline{\Delta t_{0}}=163\mu s (\sigma=15\mu s)$
for \emph{holding registers}.

\section{Discussion}
By providing a generic implementation of a Modbus/TCP driver for ROS we provide
a solution to the generalized IO problem in ROS.
While the provided driver only abstracts the workings of the communication
protocol and data interpretation (i.e. handling byte and word orders), we
believe this to be of great benefit to ROS developers. Using the driver, ROS
developers mostly stay in their domain, using high level programming languages
on general purpose computers, writing configuration files in standardized
formats and interfacing the ROS by use of ROS topics. Mapping individual IOs as
individual topics allow for great portability and flexibility while keeping the
requirements for the driver low. Adapting topic descriptors to those required by
a specific application can easily be achieved using ROSs' remapping methods;
thus allowing simple code and application component reuse. Aggregating data from
multiple individual IOs to higher level message/data types, conveying semantic
information, we consider trivial in the instant all data is available to a
developer via ROS topics.

Even though Modbus/TCP may seem like an outdated choice and solution for the
problem at hand, it is the most widespread protocol understood by suitable
hardware devices, when compared to OPC UA and MQTT as the identified
competitors.
Modbus/TCPs' simplicity results in good compatibility with a number of devices
and allows good horizontal scalability of the solution across devices.
With IoT cloud vendors supporting Modbus/TCP (compare \cref{drivers}), we do
even expect to see an increase in Modbus/TCP implementations for low cost
devices. The simultaneously good support for Modbus/TCP in industrial
applications and good availability of bus couplers and compatible PLCs ensures
applicability of our solution in industrial applications. This support, combined
with the wealth of available IO modules, further ensures a high flexibility in
IO choices and rapid prototyping of applications.

The performance evaluation shows the introduction of a time delay $<0.5ms$ for
the first value of a Modbus/TCP read-request to be available on the ROS network,
with subsequent readings of a request being available after $<0.2ms$
consistently. In our experience, these values are at least an order of magnitude
lower than the response times of actual Modbus/TCP slaves. The performance
penalty introduced by the driver is thus assumed to be negligible in ROS applications.

We have successfully used our solution with multiple \emph{Wago} and
\emph{Beckhoff} PLCs and bus couplers, each with a range of different IO
modules. The created IO capabilities have been used to control and read e.g. a
robotic gripper, switches and high resolution distance sensors. The integration
of each new device being in the range of minutes.

\section{Conclusion}

We have delivered the concept and necessary implementation of a solution to the
generalized problem of hardware IO integration into ROS systems. All while
delivering a solid rationale for the choice of Modbus/TCP as the underlying
technology.

As OPC UA and the combination of MQTT and Sparkplug do provide significant
advantages over Modbus/TCP, we expect these technologies to replace a Modbus/TCP
based implementation in the future. Building on these technologies allows to
reduce configuration to an absolute minimum by leveraging the information model
provided by the server in case of OPC UA and the metrics definition of Sparkplug
birth certificates respectively.


The main value of this work is seen in the simplicity of the provided
tools, and in presenting and delivering the idea, of using standardized
industrial protocols to the ROS community; for researchers and hobbyists
alike. Being aware of the options available to solve a particular
problem and receiving minimal support in terms of tooling, has a great
potential for keeping individuals from rolling custom communication
protocols over non-scalable communication channels (e.g. using USB/UART
and platforms like Arduino as IO hardware), and thus creating solutions
adhering to the core idea of ROS: Creating reusable solutions to recurring
problems.

\subsection*{Future Work}
Future work is expected in the optimization of read and write processes,
including batch writing of near-simultaneously published output data, and
reducing traffic and load on the Modbus/TCP slave by building dynamic readers
and querying subscribed topics only. Further, a notable improvement of the
configuration process may be achieved by interpreting the configuration files of
Modbus slave simulation and testing tools (e.g. WinModbus\cite{winmodbus} or
UnSlave\cite{unslave}).




\ifCLASSOPTIONcaptionsoff
  \newpage
\fi

\bibliographystyle{IEEEtran}
\IfFileExists{./bib/bibliography_final.bib}{%
  \typeout{using: bibliography_final.bib}%
  \bibliography{IEEEabrv,./bib/bibliography_final}%
  }{%
    \IfFileExists{./bib/bibliography.bib}{%
        \typeout{using: bibliography.bib}%
        \bibliography{IEEEabrv,./bib/bibliography}%
      }{%
				\typeout{bibliography not found, using empty!}%
				\bibliography{IEEEabrv,./bib/empty}%
			}%
	}%

\end{document}